\documentclass{article}
\usepackage{ijcai24}

\usepackage{times}
\usepackage{soul}
\usepackage{url}
\usepackage[hidelinks]{hyperref}
\usepackage[utf8]{inputenc}
\usepackage[small]{caption}
\usepackage{graphicx}
\usepackage{amsmath}
\usepackage{amsthm}
\usepackage{booktabs}
\usepackage{algorithm}
\usepackage{algorithmic}
\usepackage[switch]{lineno}
\usepackage{color}
\usepackage{multirow}
\usepackage{hyperref}
\usepackage{tabularx}
\usepackage{pifont}
\usepackage{amssymb}
\hypersetup{hypertex=true,
colorlinks=true,
linkcolor=black,
anchorcolor=black,
citecolor=black}

\title{Label-Guided Prompt for Multi-label Few-shot Aspect Category Detection}

\author{
    ChaoFeng Guan$^1$
    \and
    YaoHui Zhu$^2$\and
    Yu Bai$^{3}$\And
    LingYun Wang$^4$
    \affiliations
    $^{1,3,4}$Shenyang Aerospace University\\
    $^2$Beijing Normal University\\
    \emails
    chaofengya@163.com,
    yaohui.zhu@bnu.edu.cn,
}
\begin{document}

\maketitle

\begin{abstract}
Multi-label few-shot aspect category detection aims at identifying multiple aspect categories from sentences with a limited number of training instances. The representation of sentences and categories is a key issue in this task. Most of current methods extract keywords for the sentence representations and the category representations. Sentences often contain many category-independent words, which leads to suboptimal performance of keyword-based methods. Instead of directly extracting keywords, we propose a label-guided prompt method to represent sentences and categories. To be specific, we design label-specific prompts to represent sentences by combining crucial contextual and semantic information. Further, the label is introduced into a prompt to obtain category descriptions by utilizing a large language model. This kind of category descriptions contain the characteristics of the aspect categories, guiding the construction of discriminative category prototypes. Experimental results on two public datasets show that our method outperforms current state-of-the-art methods with a 3.86\%$\sim$4.75\% improvement in the Macro-F1 score.

\end{abstract}

\section{Introduction}

Multi-label Aspect Category Detection aims at detecting multiple aspects of viewpoints mentioned in a sentence,  which is an important subtask of aspect-based sentiment analysis \cite{pontiki2016semeval}.
It is difficult to obtain a sufficient number of annotated comment sentences, for example, more expensive items usually have only a small number of comments. 
As a result, some studies focus on mining multiple aspects of viewpoints with a few comments, named Multi-label Few-shot Aspect Category Detection (MFACD) \cite{Hu_mengting_2021}, which has received increasing attention from researchers in the field of artificial intelligence.

Existing MFACD methods usually use a meta-learning strategy  on a large amount of base categories to acquire transferable knowledge, which contributes to learn effective representations of category prototypes from a small number of comment sentences. Then these category prototypes are used to detect aspect categories of test comment sentences with similarities between test sentences and category prototypes.
In the MFACD task, single comment sentence usually contains several different aspect categories at the same time (see sentence in Table \ref{tab:样例说明}).
Since the category prototype generation is based on the sentence representations, the multiple categories of a sentence result in inadequate differentiation among category prototypes.

To achieve representative category prototypes, \cite{Hu_mengting_2021} extracted category-relevant keywords for sentence representation using an attention mechanism. %to mitigate the interference of category-irrelevant words.
Nevertheless, there are a large number of category-irrelevant words in comment sentences \cite{zhao2022label} such as “a”, “the”, “my”.
These large numbers of category-irrelevant words make the method based on automatic keyword extraction ineffective, since the method possibly tends to pay attention to those irrelevant words. %which affects the filtering of category-relevant words, and leads to poor discrimination of the generated category prototypes.
In order to effectively capture category-relevant keywords, \cite{zhao2023learning,zhao2022label,wang2023few} utilize category labels to assist keyword extraction.
However, there are similarities between category labels, resulting in the keywords filtered with the label are not clearly distinguishable, thus affecting the learning of discriminative category prototypes.

% 表格1：样例说明

\begin{table}
\centering
\renewcommand{\arraystretch}{1.6}
\resizebox{1\columnwidth}{!}{
\begin{tabularx}{12cm}{c|l}

    \toprule
        Aspect Category&\centerline{Sentences}\\
        \hline
        
        \multirow{4}{2cm}{
        \centering
       \colorbox [RGB]{204,255,204}{(A) ambiance}\\
       \colorbox [RGB]{255,204,204}{(B) service}\\
       \colorbox [RGB]{204,204,255}{(C) food}\\
        }
        
        &1. The \colorbox [RGB]{204,255,204}{cozy environment}, \colorbox [RGB]{255,204,204}{efficient service}, \\
        & \ \ and \colorbox [RGB]{204,204,255}{tasty pizza} made our night delightful. \\
        &2. Enjoyed the \colorbox [RGB]{204,255,204}{modern decor}, \colorbox [RGB]{255,204,204}{quick staff response}, \\
        & \ \ and \colorbox [RGB]{204,204,255}{delicious burgers}. \\
        \hline
                
        \multirow{4}{2cm}{
                \centering
               \colorbox [RGB]{255,228,196}{(G) price}\\
               \colorbox [RGB]{193,225,193}{(H) location}\\
               \colorbox [RGB]{176,224,230}{(I) cleanliness}\\
        }
        
        &3. The \colorbox [RGB]{255,228,196}{competitive prices}, \colorbox [RGB]{193,225,193}{prime location} in the city center, \\
        & \ \ and the \colorbox [RGB]{176,224,230}{impeccable cleanliness} of the rooms were impressive. \\
        &4. Guests appreciated the \colorbox [RGB]{255,228,196}{fair room rates}, \colorbox [RGB]{176,224,230}{spotless environment}, \\
        & \ \ and the \colorbox [RGB]{193,225,193}{easy access to local attractions}. \\

    \bottomrule

\end{tabularx}}
\caption{Multiple aspects of the sample are linked together, and different colors mean different categories.}
\label{tab:样例说明}
\end{table}

Although the above approaches based on keyword representations demonstrate some effectiveness, it is still difficult to explicitly obtain keywords characterizing category prototypes from sentences.
Compared to directly extracting keywords for sentence representations, 
%it is a promising approach to consider the contextual and semantic information of the whole sentence\cite{jiang2022promptbert}.
we argue that prompt learning in the large language mode (LLM) is able to obtain semantic information of category from sentences \cite{jiang2022promptbert}.
Furthermore, introducing the label information in the prompt can better capture the viewpoint categories of sentences.

Based on the above analysis, we propose a label-guided prompt (LGP) method for MFACD by enhancing sentence representations and their category prototypes.
At the level of the sentence representation, the knowledge of the pre-training model is exploited via a prompt, and further the category label is introduced into the prompt to accurately express sentence category information.
At the level of category prototype generation, we utilize another kind of prompt for a LLM to generate category descriptions, which contain the characteristics of the aspect categories, guiding the construction of discriminative category prototypes.

The contributions are summarized below:

\begin{itemize}
    \item We propose a label-guided prompt (LGP) method to enhance the representations of sentences for  MFACD.
    \item We leverage large language model to obtain category descriptions, which can effectively guide discriminative category prototype generation.
    \item Experimental results on two public datasets show that the proposed method achieves state-of-the-art performance.
\end{itemize}

\section{Related Work}

\subsection{Multi-label Few-shot Aspect Category Detection}
Previous research of aspect category detection can be divided into two categories, unsupervised methods \cite{schouten2017supervised,su2021whitening,hai2011implicit} and supervised methods \cite{jiang2019challenge,Movahedi2019AspectCD,ghadery2019mncn}. 
These methods detect aspect categories from predefined sets, heavily relying on large amounts of labeled data to train a discriminative classifier, which fails to generalize well to novel aspect categories with only a few labeled. To this end, some researches begin to explore few-shot learning, which allows for rapid adaptation to new categories by only utilizing a small number of samples. Few-shot learning aims to address the challenges of limited data and data sparsity, and this task has received extensive attention in many fields, such as text classification \cite{rios2018few,chalkidis2019large}, image recognition \cite{zhu2020attribute,zhu2020multi}, intent detection \cite{hou2021few}, and relationship classification \cite{gao2019hybrid}. 
%Few-shot learning (FSL) allows for rapid adaptation to new categories by utilizing a small number of samples, even after training on a substantial amount of data. This approach effectively addresses the challenges of limited data and data sparsity.

Meanwhile, the work Proto-AWATT \cite{Hu_mengting_2021}  first attempted to address aspect category detection in multi-label few-shot scenarios. This approach based prototypical network utilizes the attention mechanism to alleviate the noise from irrelevant aspects. In order to filter irrelevant words more efficiently, \cite{zhao2022label,wang2023few} proposed a label-guided attention strategy. However, the presence of semantically similar category labels may lead to ineffective filtering of irrelevant words. 
%FSO\cite{zhao2023learning} extracts discriminative prototypes and customizes a dedicated query vector for each category by identifying the semantic contents of samples. 
%Taking the mean value of all the instances in the support set to calculate the prototypes seems to ignore the variations between instances\cite{wang2023few}. 
In this paper, instead of directly extracting keywords for sentence representations,  we propose a label-guided prompt (LGP) method to enhance sentence representations and their category prototypes for the MFACD task.

\subsection{Prompt Learning}
Prompt learning is the use of cue message to elicit knowledge from pre-trained language models.
The pre-trained language models contain powerful abilities, and they can improve many tasks. For example, some works \cite{reimers-gurevych-2019-sentence,gao-etal-2021-simcse} leveraged the BERT to obtain enhanced sentence representations.  
%By leveraging powerful per-trained knowledge, the sentence can be effectively represented with the prompt.
%and mapping the extracted knowledge to a target space for sentence vector representation.
%To mitigate the noise at word-level for each support set instance, the sentence-level representation method has gained increasing attention. 
Despite the success of BERT in sentence embedding, the performance of the original BERT was not satisfactory \cite{li-etal-2020-sentence}. PPT \cite{gu-etal-2022-ppt} showed that designing effective prompt statements can guide the learning of the model, and this approach can perform well with a small number of samples. Meanwhile, the work PromptBERT \cite{jiang-etal-2022-promptbert} uses a novel contrastive learning method for learning better sentence representations by introducing prompt. Inspired by the PPT and the PromptBERT, we attempt to apply prompt learning for the MFACD task.

\begin{figure*}
    \centering
    % Answer: [trim={left bottom right top},clip]
    \includegraphics[width=1.00\linewidth]{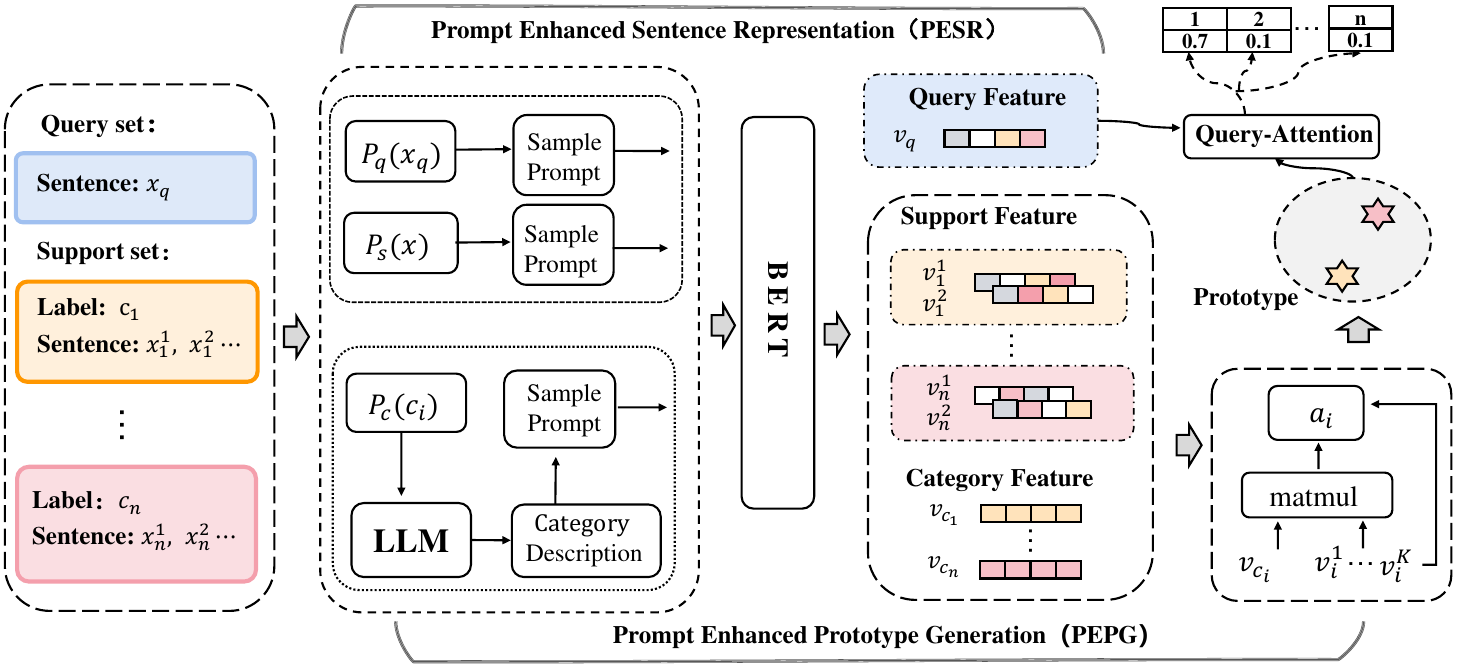}
    \caption{The overview of our proposed LGP framework. 
    %The left part describes the support set and query set samples, with different colors representing different categories.
    The left part describes the sentences of support set and query set, with different colors representing different categories.
    The middle section describes the details of PESR and PEPG. 
    PESR can enhance sentence representation by the prompt. 
    PEPG guides the prototype generation through features of category descriptions.
    }
    \label{fig:架构图}
\end{figure*}

\section{Methods}

The proposed method aims to improve the sentence representation and the category prototype via the prompt. The pipeline of the proposed method involves two crucial components Prompt Enhanced Sentence Representation (PESR) module and Prompt Enhanced Prototype Generation (PEPG) module, as depicted in Figure \ref{fig:架构图}. 
The PESR module involves learning sentence representations by using a prompt with the category label.
Based on the above improved sentence representations, PEPG module enhances the category prototype with the assistance of the category descriptions.
Before elaborating on the details of each component, we first introduce our problem definition.

\subsection{Problem Formulation}
In tackling the MFACD task \cite{snell2017prototypical}, we also adopt the meta-learning strategy, which performs on multiple $N$-way $K$-shot tasks. 
Each task consists of a support set $S$ and a query set $Q$, spanning $N$ categories with $K$ sentences per category in the support set. Namely, $S=\{x_i^j|i=1,...,N; j=1,..,K\}$, where $x_i^j$ represents the $j$-th sentence of $i$-th category. The $Q$ is composed of some query sentences $x_q$. The goal of the MFACD task is to learn a model from $S$ to recognize aspect categories of $x_q$.
%Both the support set and the query set comprise three key components: Aspect, Label, and Sample. The Aspect encompasses a variety of categories, while the Label denotes the corresponding labels for these categories. The Sample represents various categories of samples.

%Specifically, this method maps the sentence embedding vector space by learning from a support set. In this process, the embedding center calculates the prototype for each category using a small number of samples from that category in the support set. Subsequently, the predicted category of a sample is determined by measuring the distance between the query sample and these prototypes in the embedding space.

\subsection{Prompt Enhanced Sentence Representation}

Most of the current approaches use pre-trained models to obtain word representations, which are used to represent sentences.
But these approaches leads to discrepancies between the pre-trained target and the downstream task, thus limiting the performance of the pre-trained language model.
In contrast, prompt-based approaches can transform a sentence embedding task into a masked language model (MLM) task.
These methods effectively utilize the extensive knowledge of the pre-trained model and avoids the bias inherent in direct embedding \cite{brown2020language}.
Therefore, we design specific prompt templates to enhance the adaptive capacity of pre-trained model on downstream tasks.

\noindent \textbf{Prompt Template.}
We have crafted two kinds of prompt templates for sentence representations. One kind is designed for sentences of  support sets, and another is for  the query sentences.  The prompt template for sentences of support sets is as follows:
\begin{itemize}
    \item $P_s(x)$ = ``About [$x$] Category [MASK] are : [$L_x$].''
\end{itemize}
where $x$ represents a sentence of support set and $L_x$ represents the category label of $x$, and [MASK] contains $m$ learnable tokens.
The prompt template for query sentences is as follows: 
\begin{itemize}
    \item $P_q(x_q)$ = ``About [$x_q$] Category [MASK].''
\end{itemize}
where $x_q$ represents a query sentence.

%句子表示
\noindent \textbf{Sentence Representation.}
Given a sentence $x$, we can obtain the prompt with the above templates. Without loss of generality, we denote the prompt of $x$ as $P(x)$, where $P(x)$ can be $P_s(x)$ or $P_t(x)$.
%$x_{prompt}$ by splicing the input x to prompt with $P(\cdot)$ template.
Subsequently, the $P(x)$ is tokenized to fed a pre-trained model $E(\cdot)$ to obtain the hidden layer vector of the [MASK], which is denoted as $h \in R^{m \times d}$:
\begin{equation}
\begin{matrix}
h = E(P(x)) 
\end{matrix} 
\label{eq1}
\end{equation}
where $m$ is the number of tokens, and $d$ is the dimension of feature vector. Multiple tokens are incorporated into a sentence template to enhance the sentence representation. 
The final feature representation of $x$ is obtained by a mean pooling operation of $h$. The formalization is as follows:
%is $i^{th}$ category $j^{th}$ sample feature $h_i^j$, obtained through the mean pooling layer:
\begin{equation}
\begin{matrix}
v = M_p(h)
\end{matrix} 
\label{eq2}
\end{equation}
where $M_p(\cdot)$ means the mean pooling operation and $v \in R^{1\times d}$ is the sentence representation.

According to the above method of sentence representations, we can obtain sentence representations of support set, which can be denoted as $V=\{v_i^j|i=1,...,N; j=1,..,K\}$, which $v_i^j$ represents the $j$-th sentence feature representations of $i$-th category.
Similarly, the query sentence is fed to the encoder with the $P_q(\cdot)$ template to obtain its representations $v_q$.
The per-trained model effectively extract highly relevant information via the prompt for the entire sentence, obtaining precise sentence representations.

\subsection{Prompt Enhanced Prototype Generation}
To tackle the challenge of imprecise category prototypes in few-shot learning scenarios,  we harness the sophisticated capabilities of LLM for meticulous label description. These descriptions can effectively aid in the precise formation of category prototypes.

%We produce elaborate descriptions of categories through LLM with labels that act as a guide for prototype generation, 
%then we by designing a series of specific prompts, extract the characteristics and attributes of aspect categories.

\noindent \textbf{Category Descriptions.}
To produce elaborate category descriptions, we adopt
a language model prompt template, denoted as $P_c(c_i)$=``Provide a comprehensive description of [$c_i$].''
where $c_i$ symbolizes the category label. The purpose of this prompt is to distinctly encapsulate and highlight the characteristics of category. 
Subsequently, the crafted prompt $P_c(c_i)$ is fed into a LLM to generate detailed descriptions about the $c_i$. This process can be formalized as:
\begin{equation}
\begin{matrix}
D_i = G_{M}(P_c(c_i))
\end{matrix}
\end{equation}
where $D_i$ represents category descriptions and $G_{M}$ represents the large language model. Leveraging the LLM's proficiency in generating detailed descriptions, which can extract the profound semantics and core characteristics inherent to each category. These are some examples of category descriptions as follows:
\begin{enumerate}
    \item food\_food\_bread:Bread is a staple food made by baking a dough of flour and water, typically described based on its type (such as whole wheat, white bread), size (such as one loaf, one slice), texture (such as soft or hard), ingredients (such as containing kernels or seeds), and flavor (such as sweet or salty).
    \item food\_mealtype\_breakfast:Breakfast is the first meal of the day, typically consumed in the morning, consisting of food items such as cereals, eggs, toast, pancakes, fruits, and beverages like coffee or juice.
    \item food\_food\_side\_pasta:The food item in question is a side dish consisting of pasta, which is typically prepared from durum wheat dough and served as an accompaniment to a main course in various cuisines.
    \item food\_portion:Food portion refers to the specific amount of food provided or consumed during a meal, usually described by weight (such as grams, kilograms), volume (such as milliliters, liters), portion size (such as one or half portions), comparison (such as a football sized watermelon), cutlery unit (such as a bowl of rice), sensory description (such as a sumptuous dinner), or appropriate number of people (such as two servings).  
\end{enumerate}
It can be observed that these category descriptions contains a wealth of information about the category. According to Eq. (\ref{eq1}) with the prompt template $P_s(\cdot)$ and Eq. (\ref{eq2}), we can obtain the representations of $D_i$, denoted as $v_{c_i} \in R^{1\times d}$.

%Then the $D_i$ is f
%\begin{equation}
%\begin{matrix}
%h_{C_i} = E(P_s(D_i)) 
%\end{matrix} 
%\end{equation}

%\noindent where $LLMG(\cdot)$ represents a generation of LLM, the notation $h_{c_i} \in R^{m \times d}$.

%这里可以使用多个类别描述！！！补充实验
%Given that automatically generated text may contain noise or incomplete and relevant information, these descriptions are further edited and refined as needed.
%To improve their accuracy and relevance, it is important to ensure that the descriptions used in subsequent steps are highly explicit and relevant.

\noindent \textbf{Prototype Generation.}
The $v_{c_i}$ is used to provide a guidance for the prototype generation with sentence representations. 
%The feature vector $v_{c_i}$ for the $i^{th}$ category is derived using a mean pooling layer. 
%This process involves averaging the features to obtain a consolidated representation for the specified category.
%\begin{equation}
%\begin{matrix}
%v_c{_i} = MPL(h_c{_i})
%\end{matrix} 
%\end{equation}
%\noindent where $v_c{_i},h_c{_i} \in R^{1 \times d}$.
The discriminant prototype $r_i$ for the $i^{th}$ class is extracted by calculating importance based on the category descriptions and the sentence representations $V_i \in R^{K \times d}$. The $V_i$ is composed of $\{v_i^j|j=1,...,K\}$. The common used method of obtaining the prototype is mean pooling of these sentence representations. In this kind of method, the weight of each sentence is equal. We think that some sentences possibly contain a mass of noise information. The same weight for all sentences can not avoid those noise information. Thus, we use the description representations $v_{c_i}$  generated by LLM to reduce the importance of those noise sentence. 
%s_vector_rep torch.Size([1, 5, 768])
% v_{c_i} [1, 768]
%v*vc=[1, 5, 768] -> W1 ->[1, 5, 2]->mean->[1, 5]
%A_1 torch.Size([1, 5])
%proto_raw torch.Size([1, 5, 768])
%r_i torch.Size([1, 768])
The importance of sentence representations $a_i \in R^{1 \times K}$ is: 
\begin{equation}
\begin{matrix}
%a_i =\text{softmax}(M_p(v_{c_i} \odot V_{i}))
a_i =\text{softmax}(M_p(v_{c_i} * V_{i})^T)

\end{matrix} 
\end{equation}
\noindent where $\text{softmax()}$ is the softmax function and $\odot$ is the Hadamard product, i.e., element-wise multiplication. In this Hadamard product, the $v_{c_i}$ extends its size to reach $V_{i}$ by row copy.  Using the attention weight of each sentence $a_i$, we can obtain the prototype $r_i \in R^{1\times d}$ .
\begin{equation}
\begin{matrix}
r_i = a_i*V_i,
\end{matrix} 
\end{equation}
\noindent where $*$ is the matrix multiplication.  
According to the above calculation, we can obtain all category prototypes $\{r_i|i=1, 2, ..., N\}$.

\subsection{Training and Inference}

%Firstly, calculate the similarity distances between each sample in the query set and the corresponding prototype vectors. Following this, compute the loss using these distances. Based on the loss calculations, derive the predictions for the model.

\noindent \textbf{Training.}
In noisy scenarios, the query sample may contain more than one target aspect category. 
Following the previous works \cite{Hu_mengting_2021}, we employ query-attention to customize multiple prototype-specific query representations  $v_i^q \in R^{d}$, which can be obtained by:
%N=5,Q=1   
%[1, 768](q)*[5, 768](r) -> [5, 768] ->W2 -> [5,2]->[5]*[]

%A_2 torch.Size([5, 25])
%v_i^q torch.Size([25, 768])
\begin{equation}
\begin{matrix}
v_i^q = \text{softmax}(\text{tanh}(r_i) \odot v_q) \odot v_q
\end{matrix} 
\end{equation}
\noindent where $\tanh()$ is the tanh function.

The final training loss is:
\begin{equation}
\begin{matrix}
L = - \frac{1}{N} \sum\limits_{i=1}^{N} y_i (\hat{y}_i-log{\sum\limits_{i=1}^{N}e^{\hat{y}_i}})
\end{matrix} 
\end{equation}
where $y_i$ is the ground truth label of $i$-th category (i.e., $y_i$=0 or 1). $\hat{y}_i$ is calculated as follows:
\begin{equation}
\begin{matrix}
\hat{y}_i = \frac{Cos(r_i, v_i^q)-\mu}{\delta}
\end{matrix} 
\end{equation}
\noindent where $Cos(r_i, v_i^q)$ represents the cosine similarity between the prototype $r_i$ and prototype-specific query representations $v_i^q$, $\mu$ is the mean of cosine similarity and $\delta$ is the variance of cosine similarity.

\noindent \textbf{Inference.}
In multi-label classification, compared with the method of the fixed threshold, the method of the dynamic threshold adopts a adaptive value based on label distribution, accurately adapting to different label characteristics. We use a simple and effective strategy of dynamic thresholds by combining varies of statistical magnitude. This dynamic threshold $y_i^d$ of $i$-th category is calculated as follows: 
\begin{equation}
\begin{matrix}
y_i^d = \alpha \cdot m(\hat{y}_i) + \beta \cdot \sigma(\hat{y}_i) + \gamma \cdot \max(\hat{y}_i) +\\ (1 - \gamma) \cdot \min(\hat{y}_i)
\end{matrix} 
\end{equation}
\noindent where $m()$ represents the mean value, $\sigma()$ is the standard deviation, max() is the maximum and min() is the minimum value. $\alpha$, $\beta$, and $\gamma$ are adjustable parameters.

\begin{table}
    \renewcommand{\arraystretch}{1.2}
    \setlength{\tabcolsep}{3.8mm}{
    \begin{tabular}{lccc}
        \toprule
            Dataset &\#cls. &\#sent./cls.& \#sent.\\
            \hline
            FewAsp(multi) &100 &400& 40000\\
            FewAsp &100 &630 &63000\\
        \bottomrule
    \end{tabular}}
    \caption{Dataset statistics. \#cls. and \#sent. denote the number of classes and samples, respectively.}
    \label{tab:Datasets}
\end{table}

%& 5-way 5-shot & 5-way 10-shot & 10-way 5-shot & 10-way 10-shot \\
%& AUC   F1 & AUC   F1 & AUC   F1 & AUC   F1 \\
%\multicolumn{2}{c}{\textbf{Support set}}\\ 
%            \hline
%            Aspect Category&\centerline{Sentences}\\
%            \hline
%            \multirow{2}*{(A)food\_seafood\_fish}&1.Quick service, fish is fresh and delicious.\\
%            &2.Short wait, excellent service, top-notch fish quality.\\

\begin{table*}[ht]
    \centering
    \renewcommand{\arraystretch}{1.3}
    \setlength{\tabcolsep}{2mm}{
    \begin{tabular}{lcccc}
        \toprule
        \multirow{2}*{Models}& \underline{5-way 5-shot} & \underline{5-way 10-shot} 
                            & \underline{10-way 5-shot} & \underline{10-way 10-shot} \\
                    & AUC \qquad  F1 & AUC \qquad  F1 & AUC \qquad  F1 & AUC  \qquad F1 \\
        \hline
        Prototypical 
        Network \cite{snell2017prototypical}     & 88.88 \quad 66.96& 91.77 \quad 73.27& 87.35 \quad 52.06& 90.13 \quad 59.03\\
        IMP\cite{allen2019infinite}         & 89.95 \quad 68.96& 92.30 \quad 74.13& 88.50 \quad 54.14& 90.81 \quad 59.84\\
        Proto-HATT \cite{gao2019hybrid}  & 91.54 \quad 70.26 &93.43 \quad 75.24& 90.63 \quad 57.26& 92.86 \quad 61.51\\
        Proto-AWATT \cite{Hu_mengting_2021} & 93.35 \quad 75.37 &95.28 \quad 80.16& 92.06 \quad 65.65& 93.42 \quad 69.70\\
                LDF \cite{zhao2022label} & 94.65 \quad 78.27 &95.71 \quad 81.87& 92.74 \quad 67.13& 94.29 \quad 71.97\\
                LPN \cite{LiuLPN} & 96.45 \quad 82.22 &97.15 \quad 84.90& 95.36 \quad 71.42& 96.55 \quad 76.51\\
                FSO \cite{zhao2023learning} & 96.92 \quad 83.44 &97.38 \quad 85.08& 95.65 \quad 73.78& 96.28 \quad 76.58\\
        LGP(Ours)   & \textbf{97.37 \quad 87.49} &\textbf{97.49 \quad 87.67}&\textbf{96.33 \quad 77.92}& \textbf{96.69 \quad 78.95}\\
        \bottomrule
    \end{tabular}}
    \caption{Comparisons of AUC (\%) and Macro-F1 (\%) score on FewAsp. The best score is in bold.}
    \label{tab:FewAsp}
\end{table*}

\begin{table*}[ht]
    \centering
    \renewcommand{\arraystretch}{1.3}
    \setlength{\tabcolsep}{2mm}{
    \begin{tabular}{lcccc}
        \toprule
        \multirow{2}*{Models}& \underline{5-way 5-shot} & \underline{5-way 10-shot} 
                            & \underline{10-way 5-shot} & \underline{10-way 10-shot} \\
                    & AUC \qquad  F1 & AUC \qquad  F1 & AUC \qquad  F1 & AUC  \qquad F1 \\
        \hline
        Prototypical 
        Network \cite{snell2017prototypical}     & 89.67 \quad 67.88& 91.60 \quad 72.32& 88.01 \quad 52.72& 90.68 \quad 58.92\\
        IMP \cite{allen2019infinite}         & 90.12 \quad 68.86& 92.29 \quad 73.51& 88.71 \quad 53.96& 91.10 \quad 59.86\\
        Proto-HATT \cite{gao2019hybrid}  & 91.10 \quad 69.15 &93.03 \quad 73.91& 90.44 \quad 55.34& 92.38 \quad 60.21\\
        Proto-AWATT \cite{Hu_mengting_2021} & 91.45 \quad 71.72 &93.89 \quad 77.19& 89.80 \quad 58.89& 92.34 \quad 66.76\\
                LDF \cite{zhao2022label} & 92.62 \quad 73.38 &94.34 \quad 78.81& 90.87 \quad 62.06& 92.93 \quad 68.23\\
                LPN\cite{LiuLPN} & 95.66 \quad 79.48 &96.55 \quad 82.81& 94.51 \quad 67.28& 95.66 \quad 71.87\\
                FSO \cite{zhao2023learning} & 96.01 \quad 81.04 &96.67 \quad 82.22& 94.93 \quad 70.26& 95.71 \quad 72.46\\
        LGP(Ours)   & \textbf{97.67 \quad 85.22} &\textbf{97.86 \quad 86.08}& \textbf{95.89 \quad 75.01}& \textbf{96.35 \quad 76.97}\\
        \bottomrule
    \end{tabular}}
    \caption{Comparisons of AUC (\%) and Macro-F1 (\%) score on FewAsp(multi). The best score is in bold.}
    \label{tab:FewAsp(multi)}
\end{table*}

%此实验为 5/10-way、2/3-shot的实验
\begin{table*}[ht]
\centering
\renewcommand{\arraystretch}{1.3}
\setlength{\tabcolsep}{5mm}{
\begin{tabular}{lcccc}
    \toprule
    \multirow{2}*{Models}& \underline{5-way 2-shot} & \underline{5-way 3-shot} 
    & \underline{10-way 2-shot} & \underline{10-way 3-shot} \\
    & AUC \qquad  F1 & AUC \qquad  F1 & AUC \qquad  F1 & AUC  \qquad F1 \\
    \hline

    LDF \cite{zhao2022label} & 91.30 \quad 70.57 &91.23 \quad  70.51&  89.69 \quad 57.45& 90.16 \quad  59.67\\
    LPN \cite{LiuLPN} & 92.90 \quad 70.69 &94.21 \quad 74.55& 91.36 \quad 56.11&  92.67 \quad  60.34\\
     FSO \cite{zhao2023learning} & 95.14 \quad 77.44 & 95.27 \quad 78.58& 93.97 \quad  66.41& 94.32 \quad  68.28\\
    LGP(Ours)   & \textbf{95.44 \quad 81.47} &\textbf{96.15 \quad 83.60}& \textbf{94.50 \quad 71.72}& \textbf{94.89 \quad 72.98}\\
    \bottomrule
\end{tabular}}
\caption{Comparison of AUC (\%) and Macro-F1 (\%) score on FewAsp(multi). The best score is in bold.}
\label{tab:5/10-way 2/3-shot}
\end{table*}

\begin{figure*}
    \centering
    % Answer: [trim={left bottom right top},clip]
    \includegraphics[width=1\linewidth]{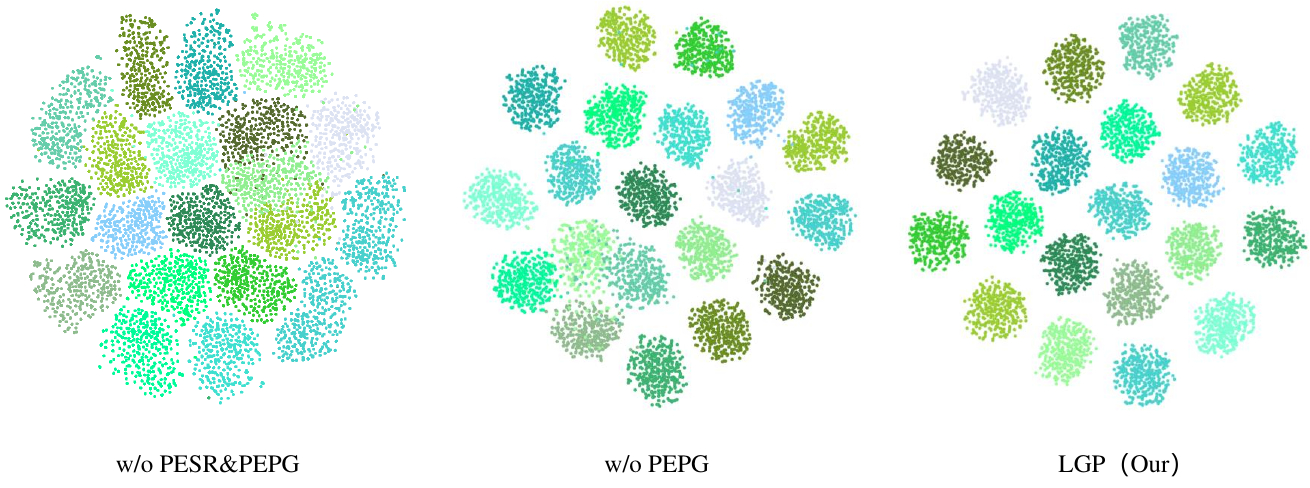}
    %trim={2.4cm 7.5cm 3cm 2.5cm},clip]
    
    \caption{Feature visualization of category prototype on FewAsp(Multi). The category prototypes are calculated from 3000 5-way 5-shot tasks. Different colors means different categories.
    %where each label is distinguished by a different color. we randomly sampled 3000 tasks of 5-way 5-shot from the FewAsp (multi) test set to visualize each sample vector.
    }
    \label{fig:Visualization}
\end{figure*}

\section{Experiments}

\subsection{Datasets and Experimental Setups}

\noindent \textbf{Datasets.}
%In order to verify the validity of LGP model, two publicly available datasets \cite{Hu_mengting_2021} are used in this paper for experimental, these datasets are constructed from Yelp\_aspect aspect \cite{Bauman_Liu_Tuzhilin_2017} with few-shot aspect category detection datasets, dataset is shown in Table \ref{tab:Datasets}.
In order to verify the validity of LGP model, two publicly available datasets FewAsp (multi) and FewAsp form \cite{Hu_mengting_2021} are used for experimental evaluation.
The FewAsp(multi) dataset consists of sentences featuring multiple aspect categories.
The FewAsp dataset comprises sentences with mixed (one or more) category types.
The above two datasets have the same 100 aspect categories, of which 64 aspect categories are used for training, 16 aspect categories for validation, and 20 aspect categories for testing.
General information of two datasets is presented in Table \ref{tab:Datasets}.

%FewAsp(multi) and FewAsp: these datasets are constructed from Yelp\_aspect aspect \cite{Bauman_Liu_Tuzhilin_2017} with few-shot aspect category detection datasets, each having the same 100 aspect categories, of which 64 aspect categories are used for training, 16 aspect categories for validation, and 20 aspect categories for testing.
%The FewAsp(multi) dataset consists of sentences featuring multiple aspect categories, whereas the FewAsp dataset comprises sentences with mixed (one or more) category types. 

\noindent \textbf{Implementation Details.}
In our study, we employ BERT-uncased English version \cite{Devlin2019bert} as the encoder, with the implementation in PyTorch. 
The LLM GPT-3.5-Turbo is chosen for generating category descriptions.
All experiments are conducted on a single NVIDIA A100, utilizing CUDA version 12.2. 
During training, we use the AdamW optimer \cite{loshchilov2017decoupled} with a learning rate of 5e-5.
And we set the hyper-parameters $\alpha$, $\beta$ and $\sigma$ with 0.3, 0.7 and 0.7, respectively.
During each epoch, 800 tasks are randomly sampled for training. In each task, $N\in\{5,10\}$, $K\in\{2,3,5,10\}$, and the number of query instances of each category is 5.

%the results of these tests were taken as the mean value as the experimental results. 

%In each dataset, we construct four MFACD tasks, where N=5,10 and K=5, 10. Then under the same conditions, we set N=5, 10 and K=2, 3 to further analyze the performance in fewer sample scenarios.
%And the number of query instances each category is 5.

\noindent \textbf{Evaluation Metric.}
Following \cite{Hu_mengting_2021}, we use Macro-F1 and AUC scores as the evaluation metrics.
600 tasks sampled from test data are used for evaluation, and the average performance on the 600 tasks is reported for comparisons.
%The commonly used 5-way and 10-way settings are adopted.
%Following \cite{Hu_mengting_2021}. 
%This task can be considered as a multi-label classification task with AUC (Area Under Curve) and macro F1 as evaluation metrics.

\subsection{Experimental Results and Analysis}

%In this paper, the research methodology is compared with other baseline models, FewAsp(multi) and FewAsp datasets using 5/10-way and 5/10-shot experimental conditions were compared.

%The experimental results are summarized in Tables \ref{tab:FewAsp}, \ref{tab:FewAsp(multi)} and \ref{tab:5/10-way 2/3-shot}, with the best score for each metric in bold.

%\noindent \textbf{Analysis of results.}
%We delve into the data presented in Tables \ref{tab:FewAsp} and \ref{tab:FewAsp(multi)}. For the 5-way 5-shot scenario, we have modified the [MASK] position to include 10 learnable tokens.
%In this study, we contrast the research methodology with other foundational models using the FewAsp and FewAsp(multi) datasets.

%This paper presents a comparative analysis of the research methodology against baseline models using the FewAsp(multi) and FewAsp datasets under 5/10-way and 5/10-shot experimental setups.

%To gain a more nuanced understanding of performance in fewer scenarios,  Our experimental setup of 5/10 way configuration and 2/3 gun was verified on FewAsp(multi). The results of these detailed studies are presented in the accompanying in Table \ref{tab:5/10-way 2/3-shot}.

%FewAsp
The comparative experimental results on the FewAsp dataset are shown in Table \ref{tab:FewAsp}. It can be observed that the proposed LGP demonstrates the superior performance than the existing methods under all experimental evaluations. 
It is worth noting that LGP surpass current state-of-the-art method FSO \cite{zhao2023learning} with 4.05\% and 4.14\% gains on F1 score under 5-way 5-shot setting and 10-way 5-shot setting, respectively.  

%The performance of LGP not only shows its efficiency and stability in handling few-shot learning tasks but also proves its exceptional ability in balancing precision and recall. 

% FewAsp(multi)
The comparative results on the FewAsp(multi) dataset are shown in Table \ref{tab:FewAsp(multi)}. We also observe that the proposed LGP achieves state-of-the-art performance under all experimental evaluations.
Compared with FSO, the LGP obtains large performance gains on F1 score with 4.08\%, 3.86\%, 4.75\% and 4.51\% improvements under 5-way 5-shot, 5-way 10-shot, 10-way 5-shot and 10-way 10-shot settings, respectively. The LGP obtains 1.66\%, 1.21\%, 0.96\% and 0.64\% improvements on AUC over FSO, under 5-way 5-shot, 5-way 10-shot, 10-way 5-shot and 10-way 10-shot settings, respectively.

In addition, following the work FSO \cite{zhao2023learning}, we also conduct experiments on FewAsp(multi) with fewer training samples such as 2-shot and 3-shot. The experimental results are exhibited in Table  \ref{tab:5/10-way 2/3-shot}.
The proposed method LGP also achieves the best performance under all experimental settings. And the relatively large improvements on F1 score also can be found.
%Across all experimental setups, including 5-way 2-shot, 10-way 2-shot, and for different shot numbers, LGP consistently achieved the highest AUC and F1 scores. 
%Specifically, in the 5-way 2-shot condition, LGP attained an AUC of 95.44\% and an F1 score of 81.47\%, while in the 10-way 3-shot condition, it reached an AUC of 94.89\% and an F1 score of 72.98\%. These results significantly surpassed other models like FSO and LPN. For instance, FSO's best performance in the 10-way 3-shot condition yielded an AUC of 94.32\% and an F1 score of 68.28\%. LGP's outstanding performance, especially in terms of the F1 score in these highly constrained data environments, highlights its robust capability in balancing precision and recall. 

%Overall, the proposed LGP excels in few-shot classification, accurately classifying and recalling categories even under tough conditions. Its robustness and adaptability mark it as a top contender in few-shot category detection.
%These findings strongly validate LGP's effectiveness and adaptability in few-shot category detection.

To summarize, the proposed method LGP shows performance advantage on the above two datasets under different experimental settings, exhibiting the effectiveness and the superiority of the proposed method. It seems to suggest that enhancing representations of both sentences and category prototype with the prompt is a promising method to address the MFACD task.

%\noindent \textbf{Ablation Study Analysis.}
%To concretely ascertain the efficacy of our experimental methodology, we conducted comparative experiments focusing on different modules and visualization of prototype vectors.

%\multirow{2}*{\textbf{Models}}& \underline{\textbf{5-way 5-shot}} & \underline{\textbf{5-way 10-shot}} 
%                    & \underline{\textbf{10-way 5-shot}} & \underline{\textbf{10-way 10-shot}} \\
%            & AUC \qquad  F1 & AUC \qquad  F1 & AUC \qquad  F1 & AUC  \qquad F1 \\
%\hline
%\textbf{LGP(Ours)}   & \textbf{96.98 \quad 84.96} &\textbf{97.86 \quad 85.93}& \textbf{96.23 \quad 75.61}& %\textbf{96.39 \quad 77.15}\\
%\scriptsize
%\footnotesize
%\small
%\normalsize
%\large
%In the ablation study for the 5-way 5-shot scenario on FewAsp(multi), the full model with both Prompt Enhanced for Sentence Representation (PESR) and Prompt Enhanced Prototype Generation (PEPG) modules achieves a high AUC of 97.67\% and an F1 score of 85.22\%.
%我们将两个模板进行了对比可以很清楚的发现去掉两个模板都会导致结果下降，并且PESR比PEPG下降的更多，表明句子表示的方法对实验影响更大。去掉模块下降的原因可能有以下几点：

\subsection{Ablation Study}

\noindent \textbf{Effectiveness of Components.} 
The proposed method LGP mainly involves two components Prompt Enhanced Sentence Representation (PESR) and Prompt Enhanced Prototype Generation (PEPG). To verify the effectiveness of each component, we conduct ablation study, and experimental results of are shown in Table \ref{tab:Ablation}.
%The reasons for the observed decrease when the modules are removed could be attributed to several factors.
The removal of the PESR module results in a 2.18\% decrease in F1 score and a 1.66\% decrease in AUC score. This seems to mean that the absence of PESR undermines the model's comprehension of sentence context.
%and accuracy in categorization. %This suggests that the absence of PESR impairs the model's performance.
Excluding the PEPG module leads to a 1.49\% reduction in F1 scores and a 1.54\% reduction in AUC scores, indicating the vital role of PEPG in forming representative category prototypes.
The combined removal of both PESR and PEPG causes the most substantial decline in performance, which can explain that their synergy is crucial to the model.

\begin{table}
    \renewcommand{\arraystretch}{1.2}
    \setlength{\tabcolsep}{3mm}{
    \begin{tabular}{cccccc}
        \toprule
        PESR & PEPG & AUC & $\delta$ AUC & F1 & $\delta$ F1 \\
        \hline
        \checkmark&\checkmark &\textbf{97.67}& &\textbf{85.22}&  \\
        \ding{55} &\checkmark &96.01 &-1.66 &83.04 &-2.18 \\
        \checkmark&\ding{55}   &96.13 &-1.54 &83.73 &-1.49 \\
        \ding{55}&\ding{55}  &95.09 &-2.58 &80.29 &-4.93 \\
        \bottomrule
    \end{tabular}}
    \caption{
    Ablation study on FewAsp(multi) under 5-way 5-shot setting. PESR means Prompt Enhanced Sentence Representation and PESR means Prompt Enhanced Prototype Generation.}
    \label{tab:Ablation}
\end{table}
Moreover, comparing PESR and PEPG, we observe that the performance reduction is more obvious when the  PESR is removed. This indicates the PESR is a more effective component than the PEPG and the method of sentence representation exerts a greater influence than the prototype generation. 

%1. **PESR's Impact on Sentence Representation**: PESR significantly enhances sentence context comprehension, leading to more accurate categorization, with its removal causing a marked decrease in model performance.

%2. **PEPG's Role in Prototype Generation**: PEPG is instrumental in creating distinct, representative category prototypes, and its absence results in less effective category differentiation.

%3. **Synergistic Effect**: The combined removal of PESR and PEPG leads to the most pronounced performance decline, highlighting their synergistic contribution to the model's accuracy and efficiency.

%4. **Crucial in Few-Shot Learning**: Both PESR and PEPG are critical in the few-shot learning context, where their contributions are vital for maximizing the limited available data, thereby enhancing the model's overall effectiveness.

\noindent \textbf{Visualization.}
%To enhance our analysis of performance, 
We visualize feature representations of category prototypes with and without PESR and PEPG. 
The tool of t-SNE \cite{van2008visualizing} is used to visualize the feature vectors, and the visualization results of 20 test categories on FewAsp(Multi) are illustrated in Figure \ref{fig:Visualization}.
%clearly demonstrate the prototype embedding in the feature space for the 20 categories in the test set. 
%The first graph, w/o PER\&PEG shows the situation without the use of PESR and PEPG modules, and it can be seen that the distribution of prototypes is rather messy and the boundaries between categories are not obvious. The second graph, w/o PEG shows only the absence of PEPG modules, with the distribution of prototypes relatively concentrated but still overlapping. The final graph, LGP shows the results of the complete use of PESR and PEPG modules, and the distribution of prototypes is more concentrated and separated, showing clearer class boundaries.

It can be seen that there are distribution differences in the feature representations of category prototypes before and after the application of the PESR and PEPG modules.
When we don't have any prompt to help for prototyping, namely without both PESR and PEPG module, the distribution of category prototypes is scattered with blurred boundaries between categories. When only PESR is applied, there is an improvement in the aggregation of prototypes, but the distinction between categories remains insufficient.  When both PESR and PEPG are applied, the category prototypes are not only more compact but also more clearly separated from each other, indicating that the PESR and PEPG modules can effectively enhance the generation of representative category prototypes.
%In comparison to prototypes without PESR and PEPG, our method successfully eliminates the negative effects of noise and extracts more representative prototypes.

\subsection{Discussions}

\noindent \textbf{Effect of Prompt Templates.}
A crucial issue for prompt-based tasks is finding the suitable template. 
This issue has been explored through two distinct types of templates: hard template and soft template, as detailed in Table \ref{tab:templates}. 
Hard templates are crafted manually, providing a structured and fixed format. 
In contrast, soft templates offer a learnable prompt structure, allowing the model to autonomously adapt and learn the most effective format.
According to \cite{lester2021power}, we compare the following soft prompt: ``$p_1$, [MASK], ...[$x$], $p_n$'', where $p_i$ means the learnable prompt.

\begin{table}
\centering
    \renewcommand{\arraystretch}{1.3}
    \setlength{\tabcolsep}{1mm}{
    \begin{tabular}{l|l|c}
        \toprule
            \multicolumn{2}{l}{Templates} &   F1  \\
            \hline
   
            Soft&$p_1$, [MASK], ...[$x$], $p_n$ &82.87 \\
            \hline

            \multirow{10}{0.7cm}{            
            Hard  
            }
            
            &In [x], the opinions about [$L_x$] are [MASK]. &\multirow{2}*{84.41} \\
            &\underline{In [x], the opinion that exists are [MASK].}  & \\
            
            &[$L_x$] The aspects of [$x$] are [MASK]. &\multirow{2}*{84.71} \\
            &\underline{The aspects of [$x$] are [MASK].}  &\\
            
            &[$L_x$] What are the aspects of [$x$] [MASK]. &\multirow{2}*{84.43} \\
            &\underline{What are the aspects of [$x$] [MASK].} &\\
            
             &This [$x$] means : [$L_x$] [MASK].&\multirow{2}*{84.44} \\
            &\underline{This [$x$] means : [MASK].}  &\\
            
             &About [$x$] Category [MASK] are : [$L_x$].&\multirow{2}*{85.22}\\
            &About [$x$] Category [MASK]. &\\

        \bottomrule
    \end{tabular}}
    \caption{Comparisons of Macro-F1 (\%) for different templates under the 5-way 5-shot setting on FewAsp(multi).}
    \label{tab:templates}
\end{table}

%This suggests that the specific formulation and structure of prompts play a crucial role in enhancing the model's capability to accurately analyze aspects and sentiments in the given dataset.

%token num
%\noindent \textbf{Effect of Tokens and Encoder.}
%为了更明确的评估影响，从prompt里取的隐藏层向量数量，编码器。我们进行来了一系列实验。
%To explicitly assess the impact of [MASK] on the experiment, we conducted a series of experiments in Figure \ref{fig:token_n}.
%To explicitly assess the impact, which the number of hidden layer vectors taken from the prompt, and take an encoder with the same conditions. 
%We conducted a series of experiments.
%We set up four experimental conditions in Figure \ref{fig:token_n}:
%By fixing the token of [MASK], or replacing the [MASK] position with a learnable token.
%The effect of bert on the experimental results was also verified by blocking the update of most of bert's parameters, and only updating the vector of learnable tokens at the [MASK] position.
%We conduct tests on the 5-way 5-shot FewAsp(multi) dataset.

Table \ref{tab:templates} showcases the impact of different prompt templates under 5-way 5-shot setting on FewAsp(multi). Among the templates, ``About [$x$] Category [MASK] are : [$L_x$]'' stands out with the highest performance with 85.22\% F1, indicating its effectiveness in guiding the model for aspect-based sentiment analysis. In comparison, the soft prompt scores lower at 82.87\% F1. The inclusion of the category label ([$L_x$]) in prompts generally improves performance, as seen in the superior results of structured prompts over their underlined variants without [$L_x$]. 
Despite different prompts have different effects on the model's performance, on the whole, a prompt-based approach generally leads to notable enhancements in the model.

%当我们采用相同条件的encoder时，。。。。

%当我们采用相同条件的token num时，。。。。

\begin{figure}
    \centering
 % Answer: [trim={left bottom right top},clip]
    \includegraphics[width=1\linewidth]{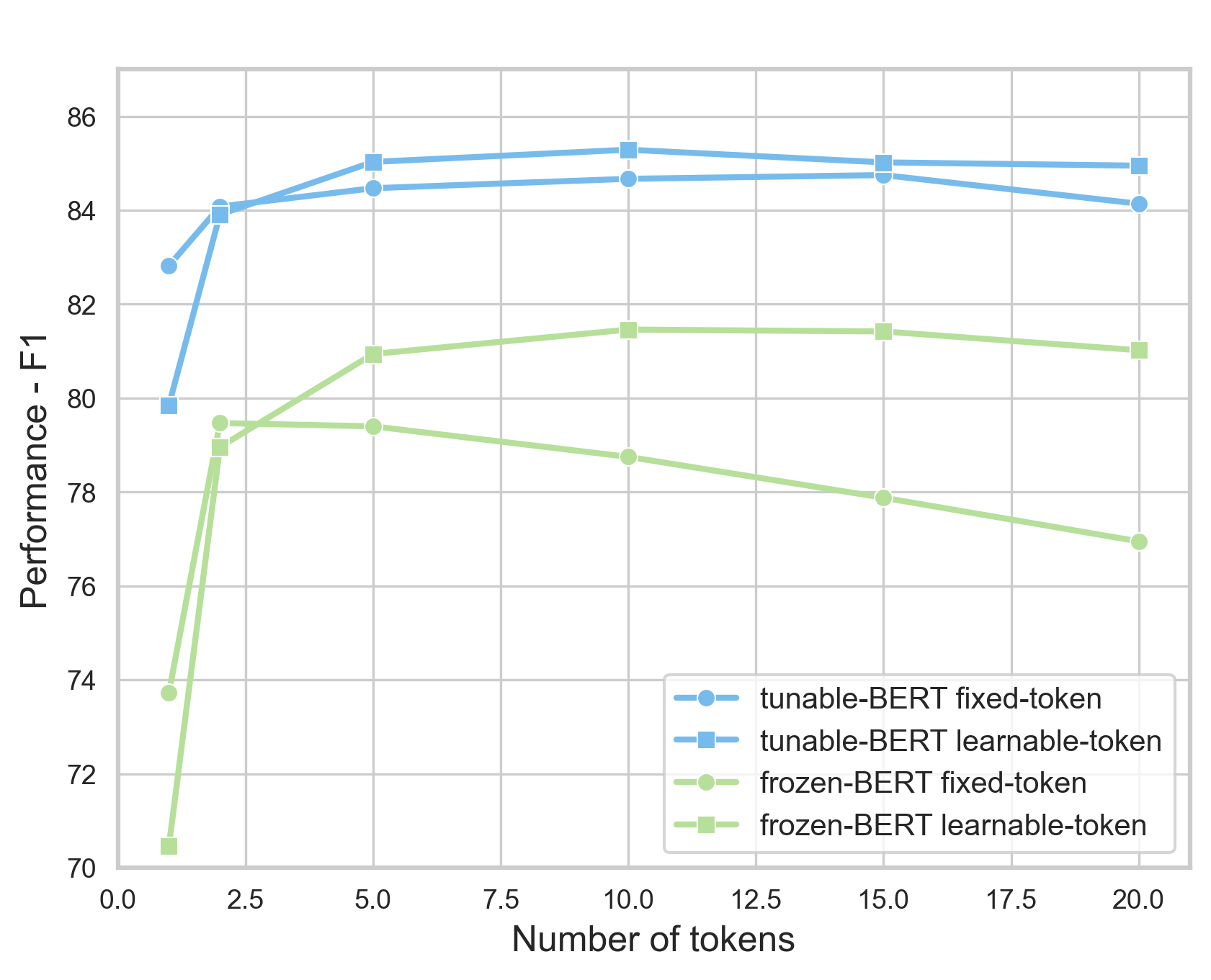}
    \caption{
    The impact on the number of tokens, token state and encoder parameters.}
    \label{fig:token_n}
\end{figure}

\noindent \textbf{Effect of Tokens.}
We explicitly assess the effect of the token from its quantity and state (i.e. fixed token or learnable token).
%By fixing the token of [MASK], or replacing the [MASK] position with a learnable token.
%As shown in Figure \ref{fig:token_n}, When the same conditions are applied to an encoder, the experimental results show that 
When the BETR is tunable (see the blue line in Figure \ref{fig:token_n}), 
the performance of both fixed token and learnable token method is improved at the beginning of the number of token increasing, after which the performance is not significantly improved. 
%adding more than one token substantially boosts the impact. Additionally, the enhancement stabilizes once the token count reaches five.
This is attributed to the improvement of the ability
with involving  multiple tokens.

When the BETR is frozen (see the green line in Figure \ref{fig:token_n}), the method of the fixed token obtain performance degrade after the number of tokens exceeds 5. However, that doesn't happen when the tokens are learnable.
It can be observed that replacing the fixed tokens  with learnable tokens contributes to the stability of our model.

\noindent \textbf{Effect of Encoder Parameters.}
To explicitly assess the effect of parameter learning in the encoder,
we explore two experiments frozen BETR (the green line in Figure \ref{fig:token_n}) and tunable BETR (the blue line in Figure \ref{fig:token_n}).
It can be observed that the proposed method with a tunable BETR is clearly superior to that with a frozen BETR under all experimental settings.  
%The effect of bert on the experimental results was also verified by blocking the update of most of bert's parameters, and only updating the vector of learnable tokens at the [MASK] position.
%Using the same number of tokens, keeping BERT's parameters fixed (except for the output layer) leads to an average reduction of 6.45\% in F1 scores.
The reason is that allowing the parameters of BERT to be trainable enables the model to dynamically adapt current MFACD tasks.
%which can be particularly advantageous in few-shot learning scenarios, potentially resulting in improved outcomes.

\section{Conclusions and Further Works}
We propose a Label-Guided Prompt (LGP) method for noisy MFACD tasks by utilizing semantic prompt to obtain better sentence representations and category prototypes.
This method employs crafted prompts integrated with label information, and it can extract vital contextual and semantic insights from the sentences.
Meanwhile, a LLM is used to generate precise category descriptions, which effectively guide the generation of more accurate category prototypes.
%especially beneficial in scenarios with limited samples.
Extensive experimental results on two datasets demonstrate that the proposed method achieves state-of-the-art performance.

In the future work, we aim to explore some novel approaches from advanced prompting mechanisms, context-sensitive learning techniques and co-evolutionary model-prompt strategies. By pushing the boundaries of aspect category detection, our research endeavors to enhance the accuracy and robustness of detecting multiple labels in situations where data availability is limited.

\bibliographystyle{named}
\bibliography{ijcai24}

\begin{thebibliography}{}

\bibitem[\protect\citeauthoryear{Allen \bgroup \em et al.\egroup }{2019}]{allen2019infinite}
Kelsey Allen, Evan Shelhamer, Hanul Shin, and Joshua Tenenbaum.
\newblock Infinite mixture prototypes for few-shot learning.
\newblock In {\em ICML}, pages 232--241, 2019.

\bibitem[\protect\citeauthoryear{Brown \bgroup \em et al.\egroup }{2020}]{brown2020language}
Tom Brown, Benjamin Mann, Nick Ryder, Melanie Subbiah, Jared~D Kaplan, Prafulla Dhariwal, Arvind Neelakantan, Pranav Shyam, Girish Sastry, Amanda Askell, et~al.
\newblock Language models are few-shot learners.
\newblock In {\em NeurIPS}, pages 1877--1901, 2020.

\bibitem[\protect\citeauthoryear{Chalkidis \bgroup \em et al.\egroup }{2019}]{chalkidis2019large}
Ilias Chalkidis, Manos Fergadiotis, Prodromos Malakasiotis, and Ion Androutsopoulos.
\newblock Large-scale multi-label text classification on eu legislation.
\newblock {\em arXiv preprint arXiv:1906.02192}, 2019.

\bibitem[\protect\citeauthoryear{Gao \bgroup \em et al.\egroup }{2019}]{gao2019hybrid}
Tianyu Gao, Xu~Han, Zhiyuan Liu, and Maosong Sun.
\newblock Hybrid attention-based prototypical networks for noisy few-shot relation classification.
\newblock In {\em AAAI}, pages 6407--6414, 2019.

\bibitem[\protect\citeauthoryear{Gao \bgroup \em et al.\egroup }{2021}]{gao-etal-2021-simcse}
Tianyu Gao, Xingcheng Yao, and Danqi Chen.
\newblock {S}im{CSE}: Simple contrastive learning of sentence embeddings.
\newblock In {\em EMNLP}, pages 6894--6910, 2021.

\bibitem[\protect\citeauthoryear{Ghadery \bgroup \em et al.\egroup }{2019}]{ghadery2019mncn}
Erfan Ghadery, Sajad Movahedi, Heshaam Faili, and Azadeh Shakery.
\newblock Mncn: A multilingual ngram-based convolutional network for aspect category detection in online reviews.
\newblock In {\em AAAI}, pages 6441--6448, 2019.

\bibitem[\protect\citeauthoryear{Gu \bgroup \em et al.\egroup }{2022}]{gu-etal-2022-ppt}
Yuxian Gu, Xu~Han, Zhiyuan Liu, and Minlie Huang.
\newblock {PPT}: Pre-trained prompt tuning for few-shot learning.
\newblock In {\em ACL}, pages 8410--8423, 2022.

\bibitem[\protect\citeauthoryear{Hai \bgroup \em et al.\egroup }{2011}]{hai2011implicit}
Zhen Hai, Kuiyu Chang, and Jung-jae Kim.
\newblock Implicit feature identification via co-occurrence association rule mining.
\newblock In {\em CICLing}, pages 393--404, 2011.

\bibitem[\protect\citeauthoryear{Hou \bgroup \em et al.\egroup }{2021}]{hou2021few}
Yutai Hou, Yongkui Lai, Yushan Wu, Wanxiang Che, and Ting Liu.
\newblock Few-shot learning for multi-label intent detection.
\newblock In {\em AAAI}, pages 13036--13044, 2021.

\bibitem[\protect\citeauthoryear{Hu \bgroup \em et al.\egroup }{2021}]{Hu_mengting_2021}
Mengting Hu, Shiwan Zhao, Honglei Guo, Chao Xue, Hang Gao, Tiegang Gao, Renhong Cheng, and Zhong Su.
\newblock Multi-label few-shot learning for aspect category detection.
\newblock In {\em ACL-IJCNLP}, pages 6330--6340, 2021.

\bibitem[\protect\citeauthoryear{Jiang \bgroup \em et al.\egroup }{2019}]{jiang2019challenge}
Qingnan Jiang, Lei Chen, Ruifeng Xu, Xiang Ao, and Min Yang.
\newblock A challenge dataset and effective models for aspect-based sentiment analysis.
\newblock In {\em EMNLP-IJCNLP}, pages 6280--6285, 2019.

\bibitem[\protect\citeauthoryear{Jiang \bgroup \em et al.\egroup }{2022a}]{jiang2022promptbert}
Ting Jiang, Jian Jiao, Shaohan Huang, Zihan Zhang, Deqing Wang, Fuzhen Zhuang, Furu Wei, Haizhen Huang, Denvy Deng, and Qi~Zhang.
\newblock Promptbert: Improving bert sentence embeddings with prompts.
\newblock In {\em EMNLP}, pages 8826--8837, 2022.

\bibitem[\protect\citeauthoryear{Jiang \bgroup \em et al.\egroup }{2022b}]{jiang-etal-2022-promptbert}
Ting Jiang, Jian Jiao, Shaohan Huang, Zihan Zhang, Deqing Wang, Fuzhen Zhuang, Furu Wei, Haizhen Huang, Denvy Deng, and Qi~Zhang.
\newblock {P}rompt{BERT}: Improving {BERT} sentence embeddings with prompts.
\newblock In {\em EMNLP}, pages 8826--8837, 2022.

\bibitem[\protect\citeauthoryear{Kenton and Toutanova}{2019}]{Devlin2019bert}
Jacob Devlin Ming-Wei~Chang Kenton and Lee~Kristina Toutanova.
\newblock Bert: Pre-training of deep bidirectional transformers for language understanding.
\newblock In {\em NAACL-HLT}, pages 4171--4186, 2019.

\bibitem[\protect\citeauthoryear{Lester \bgroup \em et al.\egroup }{2021}]{lester2021power}
Brian Lester, Rami Al-Rfou, and Noah Constant.
\newblock The power of scale for parameter-efficient prompt tuning.
\newblock {\em arXiv preprint arXiv:2104.08691}, 2021.

\bibitem[\protect\citeauthoryear{Li \bgroup \em et al.\egroup }{2020}]{li-etal-2020-sentence}
Bohan Li, Hao Zhou, Junxian He, Mingxuan Wang, Yiming Yang, and Lei Li.
\newblock On the sentence embeddings from pre-trained language models.
\newblock In {\em EMNLP}, pages 9119--9130, 2020.

\bibitem[\protect\citeauthoryear{Liu \bgroup \em et al.\egroup }{2022}]{LiuLPN}
Han Liu, Feng Zhang, Xiaotong Zhang, Siyang Zhao, Junjie Sun, Hong Yu, and Xianchao Zhang.
\newblock Label-enhanced prototypical network with contrastive learning for multi-label few-shot aspect category detection.
\newblock In {\em SIGKDD}, pages 1079--1087, 2022.

\bibitem[\protect\citeauthoryear{Loshchilov and Hutter}{2017}]{loshchilov2017decoupled}
Ilya Loshchilov and Frank Hutter.
\newblock Decoupled weight decay regularization.
\newblock {\em arXiv preprint arXiv:1711.05101}, 2017.

\bibitem[\protect\citeauthoryear{Movahedi \bgroup \em et al.\egroup }{2019}]{Movahedi2019AspectCD}
Sajad Movahedi, Erfan Ghadery, Heshaam Faili, and Azadeh Shakery.
\newblock Aspect category detection via topic-attention network.
\newblock {\em arXiv preprint arXiv:1901.01183}, 2019.

\bibitem[\protect\citeauthoryear{Pontiki \bgroup \em et al.\egroup }{2016}]{pontiki2016semeval}
Maria Pontiki, Dimitris Galanis, Haris Papageorgiou, Ion Androutsopoulos, Suresh Manandhar, Mohammed AL-Smadi, Mahmoud Al-Ayyoub, Yanyan Zhao, Bing Qin, Orph{\'e}e De~Clercq, et~al.
\newblock Semeval-2016 task 5: Aspect based sentiment analysis.
\newblock In {\em SemEval}, pages 19--30, 2016.

\bibitem[\protect\citeauthoryear{Reimers and Gurevych}{2019}]{reimers-gurevych-2019-sentence}
Nils Reimers and Iryna Gurevych.
\newblock Sentence-{BERT}: Sentence embeddings using {S}iamese {BERT}-networks.
\newblock In {\em EMNLP-IJCNLP}, pages 3982--3992, 2019.

\bibitem[\protect\citeauthoryear{Rios and Kavuluru}{2018}]{rios2018few}
Anthony Rios and Ramakanth Kavuluru.
\newblock Few-shot and zero-shot multi-label learning for structured label spaces.
\newblock In {\em EMNLP}, pages 3132--3142, 2018.

\bibitem[\protect\citeauthoryear{Schouten \bgroup \em et al.\egroup }{2017}]{schouten2017supervised}
Kim Schouten, Onne Van Der~Weijde, Flavius Frasincar, and Rommert Dekker.
\newblock Supervised and unsupervised aspect category detection for sentiment analysis with co-occurrence data.
\newblock {\em IEEE transactions on cybernetics}, 48(4):1263--1275, 2017.

\bibitem[\protect\citeauthoryear{Snell \bgroup \em et al.\egroup }{2017}]{snell2017prototypical}
Jake Snell, Kevin Swersky, and Richard Zemel.
\newblock Prototypical networks for few-shot learning.
\newblock In {\em NeurIPS}, pages 4080--4090, 2017.

\bibitem[\protect\citeauthoryear{Su \bgroup \em et al.\egroup }{2021}]{su2021whitening}
Jianlin Su, Jiarun Cao, Weijie Liu, and Yangyiwen Ou.
\newblock Whitening sentence representations for better semantics and faster retrieval.
\newblock {\em arXiv preprint arXiv:2103.15316}, 2021.

\bibitem[\protect\citeauthoryear{van~der Maaten and Hinton}{2008}]{van2008visualizing}
Laurens van~der Maaten and Geoffrey Hinton.
\newblock Visualizing data using t-sne.
\newblock {\em Journal of Machine Learning Research}, 9(86):2579--2605, 2008.

\bibitem[\protect\citeauthoryear{Wang and Iwaihara}{2023}]{wang2023few}
Zeyu Wang and Mizuho Iwaihara.
\newblock Few-shot multi-label aspect category detection utilizing prototypical network with sentence-level weighting and label augmentation.
\newblock In {\em DEXA}, pages 363--377, 2023.

\bibitem[\protect\citeauthoryear{Zhao \bgroup \em et al.\egroup }{2022}]{zhao2022label}
Fei Zhao, Yuchen Shen, Zhen Wu, and Xinyu Dai.
\newblock Label-driven denoising framework for multi-label few-shot aspect category detection.
\newblock {\em arXiv preprint arXiv:2210.04220}, 2022.

\bibitem[\protect\citeauthoryear{Zhao \bgroup \em et al.\egroup }{2023}]{zhao2023learning}
Shiman Zhao, Wei Chen, and Tengjiao Wang.
\newblock Learning few-shot sample-set operations for noisy multi-label aspect category detection.
\newblock In {\em IJCAI}, pages 5306--5313, 2023.

\bibitem[\protect\citeauthoryear{Zhu \bgroup \em et al.\egroup }{2020a}]{zhu2020multi}
Yaohui Zhu, Chenlong Liu, and Shuqiang Jiang.
\newblock Multi-attention meta learning for few-shot fine-grained image recognition.
\newblock In {\em IJCAI}, pages 1090--1096, 2020.

\bibitem[\protect\citeauthoryear{Zhu \bgroup \em et al.\egroup }{2020b}]{zhu2020attribute}
Yaohui Zhu, Weiqing Min, and Shuqiang Jiang.
\newblock Attribute-guided feature learning for few-shot image recognition.
\newblock {\em IEEE Transactions on Multimedia}, 23:1200--1209, 2020.

\end{thebibliography}

\end{document}